\documentclass[runningheads]{llncs}
\usepackage{graphicx}
\usepackage{amsmath,amssymb} 
\usepackage{color}

\usepackage{subfig}
\usepackage{multirow}
\usepackage[pagebackref=true,breaklinks=true,letterpaper=true,colorlinks,bookmarks=false]{hyperref}
\usepackage{breakcites}

\usepackage{enumitem}
\usepackage{mmstyle}

\graphicspath{{figures/}}

\begin{document}
\title{Person Search in Videos with One Portrait \\ Through Visual and Temporal Links} 

\titlerunning{Person Search Through Visual and Temporal Links}


\author{Qingqiu Huang\inst{1} \and
	Wentao Liu\inst{2,3} \and
	Dahua Lin\inst{1}}

\authorrunning{Q. Huang, W. Liu, D. Lin}

\institute{CUHK-SenseTime Joint Lab, The Chinese University of Hong Kong
\email{\{hq016,dhlin\}@ie.cuhk.edu.hk}\\
\and
Department of Computer Science and Technology, Tsinghua University \\
\and
SenseTime Research \\
\email{liuwtwinter@gmail.com}}
\maketitle
%

\begin{abstract}
In real-world applications, \emph{e.g.}~law enforcement and video retrieval,
one often needs to search a certain person in long videos with
\emph{just one portrait}.
This is much more challenging than the conventional settings for person
re-identification, as the search may need to be carried out in the environments
different from where the portrait was taken.
In this paper, we aim to tackle this challenge and propose a novel framework,
which takes into account the identity invariance along a tracklet,
thus allowing person identities to be propagated via
both the visual and the temporal links.
We also develop a novel
scheme called \emph{Progressive Propagation via Competitive Consensus},
which significantly improves the reliability of the
propagation process.
To promote the study of person search,
we construct a large-scale benchmark, which contains 127K
\emph{manually annotated} tracklets from 192 movies.
Experiments show that our approach remarkably outperforms mainstream person re-id
methods, raising the mAP from $42.16\%$ to $62.27\%$.
\footnote[1]{Code and data at http://qqhuang.cn/projects/eccv18-person-search/}
\end{abstract}


\section{Introduction}
\label{sec:intro}

\begin{figure}[t]
	\centering
	\includegraphics[width=\linewidth]{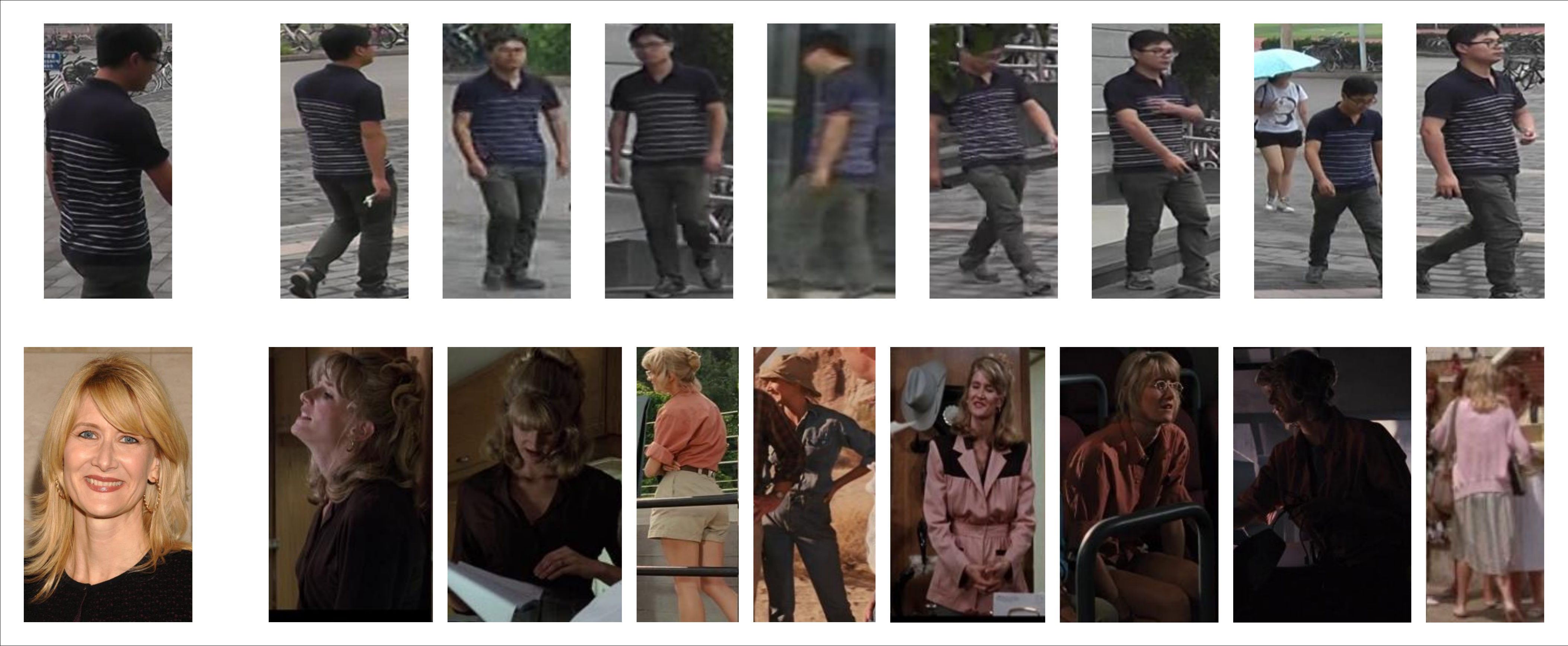}
	\caption{\small
		Person re-id differs significantly from the person search task.
		The first row shows a typical example in person re-id from the
		\emph{MARS dataset}~\cite{zheng2016mars}, where the reference and
		the targets are captured under similar conditions.
		The second row shows an example from our person search dataset
		\emph{CSM}, where the reference portrait is dramatically different
		from the targets that vary significantly in pose, clothing,
		and illumination.}
	\label{fig:teaser}
\end{figure}

Searching persons in videos is frequently needed in real-world scenarios.
To catch a wanted criminal, the police may have to go through thousands of
hours of videos collected from multiple surveillance cameras, probably with
just a single portrait.
To find the movie shots featured by a popular star, the retrieval system
has to examine many hour-long films, with just a few facial photos as the
references.
In applications like these, the reference photos are often taken in
an environment that is very different from the target environments
where the search is conducted.
As illustrated in Figure~\ref{fig:teaser}, such settings are very challenging.
Even state-of-the-art recognition techniques would find it difficult to
reliably identify all occurrences of a person, facing the dramatic variations
in pose, makeups, clothing, and illumination.

It is noteworthy that two related problems,
namely \emph{person re-identification (re-id)} and \emph{person recognition in albums},
have drawn increasing attention from the research community.
However, they are substantially different from the problem of
\emph{person search with one portrait}, which we aim to tackle in this work.
Specifically, in typical settings of person re-id~\cite{zheng2016mars,li2014deepreid,wang2016person,zheng2015scalable,hirzer2011person,gou2017dukemtmc4reid,karanam2016systematic},
the queries and the references in the gallery set are usually captured
under similar conditions, \eg~from different cameras along a street,
and within a short duration. Even though some queries can be subject to
issues like occlusion and pose changes, they can still be
identifies via other visual cues, \eg~clothing.
For person recognition in albums~\cite{zhang2015beyond}, one is typically given a diverse
collection of gallery samples, which may cover a wide range of conditions and
therefore can be directly matched to various queries.
Hence, for both problems, the references in the gallery are often
good representatives of the targets, and therefore the methods based on visual
cues can perform reasonably well~\cite{li2014deepreid,ahmed2015improved,ding2015deep,cheng2016person,xiao2016learning,zheng2016mars,zhang2015beyond,joon2015person,huang2018unifying}.
On the contrary, our task is to bridge a single portrait with
a highly diverse set of samples, which is much more challenging and
requires new techniques that go beyond visual matching.

To tackle this problem, we propose a new framework that propagates labels through
both visual and temporal links.
The basic idea is to take advantage of the \emph{identity invariance}
along a person trajectory, \ie~all person instances along a \emph{continuous}
trajectory in a video should belong to the same identity.
The connections induced by tracklets, which we refer to as the
\emph{temporal links}, are complementary to the \emph{visual links} based on
feature similarity.
For example, a trajectory can sometimes cover a wide range of
facial images that can not be easily associated based on visual similarity.
With both \emph{visual} and \emph{temporal} links incorporated, our framework
can form a large connected graph, thus allowing the identity information to be
propagated over a very diverse collection of instances.

While the combination of visual and temporal links provide a broad foundation
for identity propagation, it remains a very challenging problem to carry out
the propagation \emph{reliably} over a large real-world dataset.
As we begin with only a single portrait, a few wrong labels during propagation
can result in catastrophic errors downstream.
Actually, our empirical study shows that conventional schemes like
linear diffusion~\cite{zhu2002learning,zhou2004learning} even leads to substantially worse results.
To address this issue, we develop a novel scheme called
\emph{Progressive Propagation via Competitive Consensus}, which performs the
propagation \emph{prudently}, spreading a piece of identity information only
when there is high certainty.

To facilitate the research on this problem setting,
we construct a dataset named \emph{Cast Search in Movies (CSM)},
which contains $127K$ tracklets of $1218$ cast identities
from $192$ movies.
The identities of all the tracklets are \emph{manually annotated}.
Each cast identity also comes with a reference portrait.
The benchmark is very challenging, where the person instances for each identity
varies significantly in makeup, pose, clothing, illumination, and even age.
On this benchmark,
our approach get $63.49\%$ and $62.27\%$ mAP under two settings,
Comparing to the $53.33\%$ and $42.16\%$ mAP of the conventional visual-matching method,
it shows that only matching by visual cues
can not solve this problem well,
and our proposed framework -- \emph{Progressive Propagation via Competitive Consensus}
can significantly raise the performance.

In summary, the main contributions of this work lie in four aspects:
(1) We systematically study the problem of \emph{person search in videos},
which often arises in real-world practice, but remains widely open in research.
(2) We propose a framework,
which incorporates both the visual similarity and the identity invariance
along a tracklet, thus allowing the search to be carried out much further.
(3) We develop the \emph{Progressive Propagation via Competitive Consensus} scheme,
which significantly improves the reliability of propagation.
(4) We construct a dataset \emph{Cast Search in Movies (CSM)} with
$120K$ manually annotated tracklets to promote the study on this problem.


\section{Related Work}
\label{sec:related}

\noindent{\bf Person Re-id}.
Person re-id~\cite{zajdel2005keeping,gheissari2006person,gong2014person},
which aims to match pedestrian images (or tracklets) from different cameras within a short period,
has drawn much attention in the research community.
Many datasets~\cite{zheng2016mars,li2014deepreid,wang2016person,zheng2015scalable,hirzer2011person,gou2017dukemtmc4reid,karanam2016systematic} have been proposed to promote the research of re-id.
However, the videos are captured by just several cameras in nearby locations within a short period.
For example, the Airport~\cite{karanam2016systematic} dataset is captured in an airport
from 8 a.m. to 8 p.m. in one day.
So the instances of the same identities are usually similar enough to identify by visual appearance although with occlusion and pose changes.
Based on such characteristic of the data, most of the re-id methods
focus on how to match a query and a gallery instance by visual cues.
In earily works,
the matching process is splited into feature designing~\cite{hamdoun2008person,gray2008viewpoint,ma2012local,ma2014covariance}
and metric learning~\cite{prosser2010person,koestinger2012large,liao2015person}.
Recently,
many deep learning based methods have been proposed to jointly handle the matching problem.
\emph{Li et al.}~\cite{li2014deepreid} and \emph{Ahmed et al.}~\cite{ahmed2015improved} designed siamese-based networks which employ a binary verification loss to train the parameters.
\emph{Ding et al.}~\cite{ding2015deep} and \emph{Cheng et al.}~\cite{cheng2016person} exploit triple loss for training more discriminating feature.
\emph{Xiao et al.}~\cite{xiao2016learning} and \emph{Zheng et al.}~\cite{zheng2016mars} proposed to learn features by classifying identities.
Although the feature learning methods of re-id can be adopted for the Person Search with One Portrait problem,
they are substantially different as
the query and the gallery would have huge visual appearances gap in person search,
which would make one-to-one matching fail.

\noindent{\bf Person Recognition in Photo Album}.
Person recognition~\cite{lin2010joint,zhang2015beyond,joon2015person,li2016multi,huang2018unifying} is another related problem,
which usually focuses on the persons in photo album.
It aims to recognize the identities of the queries given a set of labeled persons in gallery.
\emph{Zhang et al.}~\cite{zhang2015beyond} proposed a Pose Invariant
Person Recognition method (PIPER), which combines three types of
visual recognizers based on ConvNets,
respectively on face, full body, and poselet-level cues.
The PIPA dataset published in~\cite{zhang2015beyond} has been widely
adopted as a standard benchmark to evaluate person recognition methods.
\emph{Oh et al.}~\cite{joon2015person} evaluated the effectiveness of
different body regions, and used a weighted combination of the scores
obtained from different regions for recognition.
\emph{Li et al.}~\cite{li2016multi} proposed a multi-level
contextual model, which integrates person-level, photo-level and group-level contexts.
But the person recognition is also quite different from the person search problem we aim to tackle in this paper,
since the samples of the same identities in query and gallery are still similar in visual appearances and the methods mostly focus on recognizing by visual cues and context.

\noindent{\bf Person Search}.
There are some works that focus on person search problem.
\emph{Xiao et al.}~\cite{xiao2017joint} proposed a person search task
which aims to search the corresponding instances in the images of the gallery without bounding box annotation.
The associated data is similar to that in re-id.
The key difference is that the bounding box is unavailable in this task.
Actually it can be seen as a task to combine pedestrian detection and person re-id.
There are some other works try to search person with different modality of data,
such as language-based~\cite{li2017person} and attribute-based~\cite{su2016deep,feris2014attribute},
which focus on the application scenarios that are different from the portrait-based problem we aim to tackle in this paper.

\noindent{\bf Label Propogation}.
Label propagation (LP)~\cite{zhu2002learning,zhou2004learning},
also known as Graph Transduction~\cite{wang2008graph,rohrbach2013transfer,sener2016learning},
is widely used as a semi-supervised learning method.
It relies on the idea of building a graph
in which nodes are data points (labeled and unlabeled) and the
edges represent similarities between points
so that labels can propagate from labeled points to unlabeled points.
Different kinds of LP-based approaches
have been proposed for face recognition~\cite{kumar2014face,zoidi2014person},
semantic segmentation~\cite{sheikh2016real}, object detection~\cite{tripathi2016detecting}, saliency detection~\cite{li2015inner} in the computer vision community.
In this paper,
We develop a novel LP-based approach called
Progressive Propagation via Competitive Consensus,
which differs from the conventional LP in two folds:
(1) propagating by competitive consensus rather than linear diffusion, and
(2) iterating in a progressive manner.


\section{Cast Search in Movies Dataset}
\label{sec:dataset}

\begin{table}[t]
	\centering
	\caption{Comparing CSM with related datasets}
	\label{tab:dataset-scale}
	\begin{tabular}{l|ccccccc}
		\hline
		Dataset   & ~~CSM~   & ~MARS\cite{zheng2016mars}   & ~iLIDS\cite{wang2016person} & ~PRID\cite{hirzer2011person}~  & ~Market\cite{zheng2015scalable} & PSD\cite{xiao2017joint}               & PIPA\cite{zhang2015beyond}        \\ \hline \hline
		task       & ~~search~  & re-id  & re-id & re-id & re-id  & det.+re-id & recog. \\ \hline
		type       & ~~video~  & video  & video & video & image  & image & image \\ \hline
		identities & ~~1,218~   & 1,261  & 300   & 200   & 1,501  & 8,432             & 2,356       \\ \hline
		tracklets  & ~~127K~ & 20K & 600   & 400   & -      & -                 & -           \\ \hline
		instances     & ~~11M~ & 1M & 44K   & 40K   & 32K    & 96K               & 63K         \\ \hline
	\end{tabular}
\end{table}

\begin{figure}[t]
	\centering
	\includegraphics[width=\linewidth]{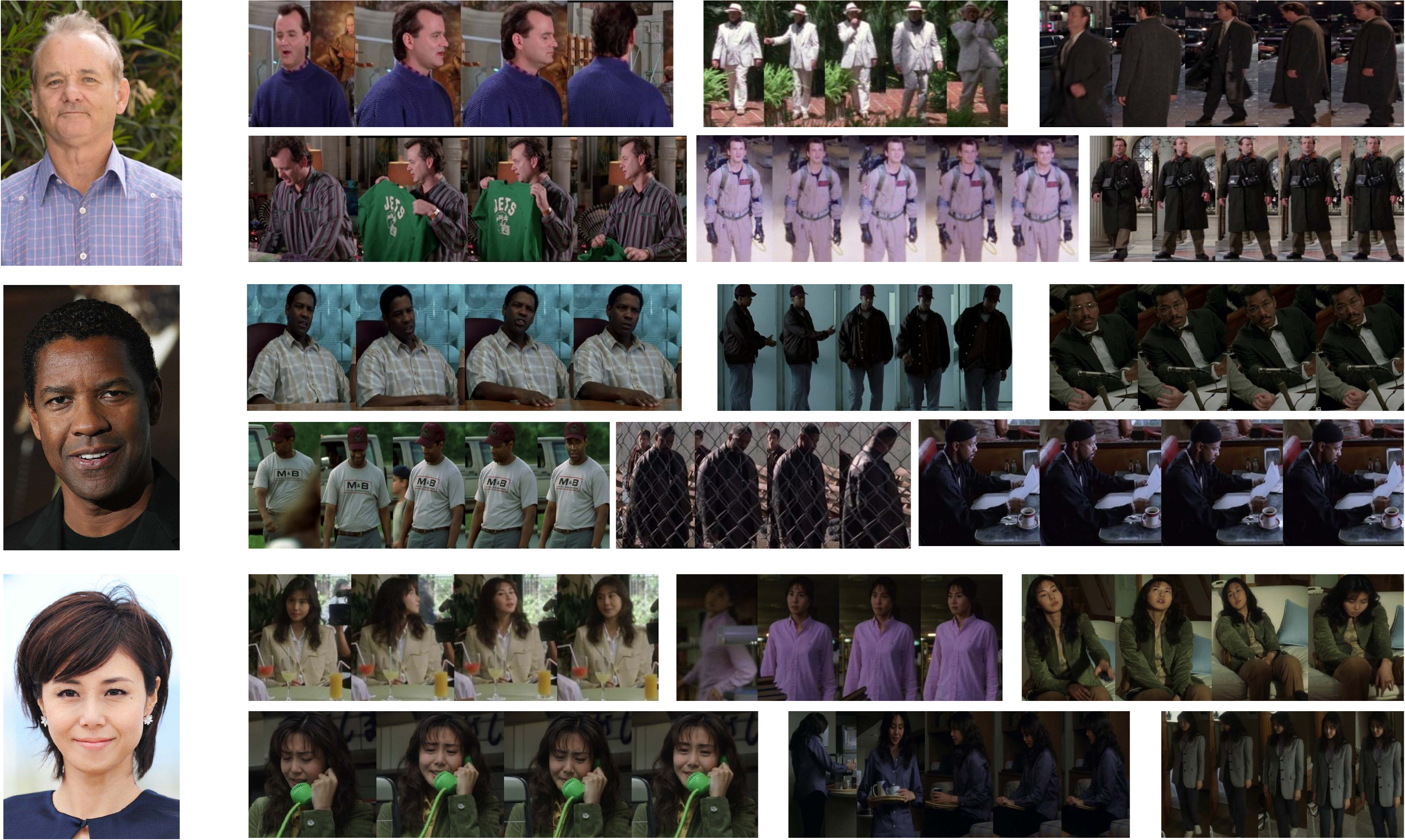}
	\caption{
		Examples of \emph{CSM} Dataset.
		In each row, the photo on the left is the query portrait and
		the following tracklets of are groud-truth tracklets of them in the gallery.
	}
	\label{fig:dataset-example}
\end{figure}

Whereas there have been a number of public datasets for person re-id~\cite{zheng2016mars,li2014deepreid,wang2016person,zheng2015scalable,hirzer2011person,gou2017dukemtmc4reid,karanam2016systematic}
and album-based person recognition~\cite{zhang2015beyond}.
But dataset for our task, namely person search with a single portrait,
remains lacking.
In this work, we constructed a large-scale dataset \emph{Cast Search in Movies (CSM)}
for this task. \emph{CSM} comprises a \emph{query set} that contains
the portraits for $1,218$ cast (the actors and actresses) and
a \emph{gallery set} that contains $127K$ tracklets (with $11M$ person instances)
extracted from $192$ movies.

We compare \emph{CSM} with other datasets for person re-id and person recognition
in Tabel~\ref{tab:dataset-scale}.
We can see that CSM is significantly larger,
$6$ times for tracklets and $11$ times more instances than
MARS~\cite{zheng2016mars}, which is the largest dataset for person re-id
to our knowledge.
Moreover, CSM has a much wider range of
tracklet durations (from $1$ to $4686$ frames) and
instance sizes (from $23$ to $557$ pixels in height).
Figure~\ref{fig:dataset-example} shows several example tracklets
as well as their corresponding portraits, which are very diverse
in pose, illumination, and wearings.
It can be seen that the task is very challenging.

\begin{figure}[t]
	\centering
	\subfloat[\label{fig:statistic-1}]{\includegraphics[width=0.28\linewidth]{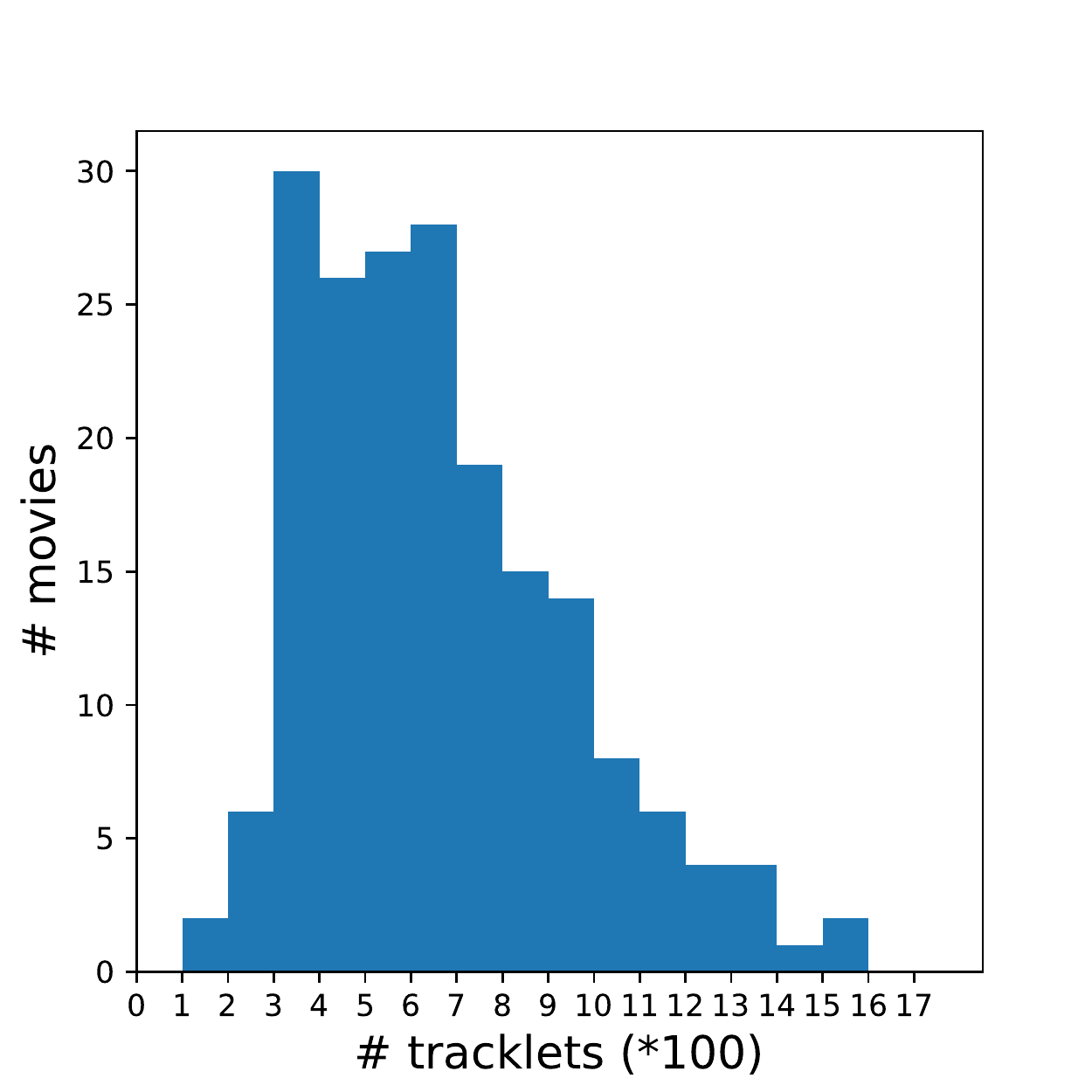}} \hfill
	\subfloat[\label{fig:statistic-2}]{\includegraphics[width=0.67\linewidth]{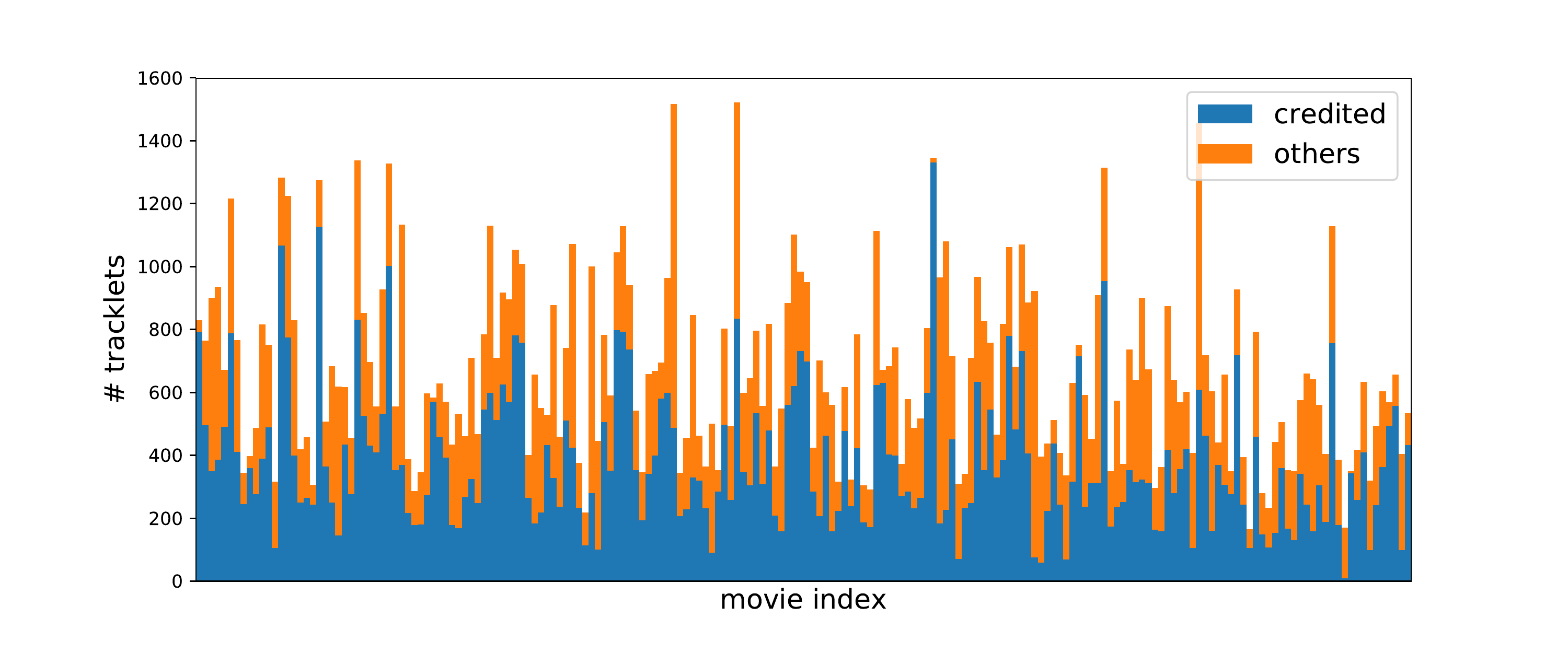}} \\ \vspace{-5pt}
	\subfloat[\label{fig:statistic-3}]{\includegraphics[width=0.32\linewidth]{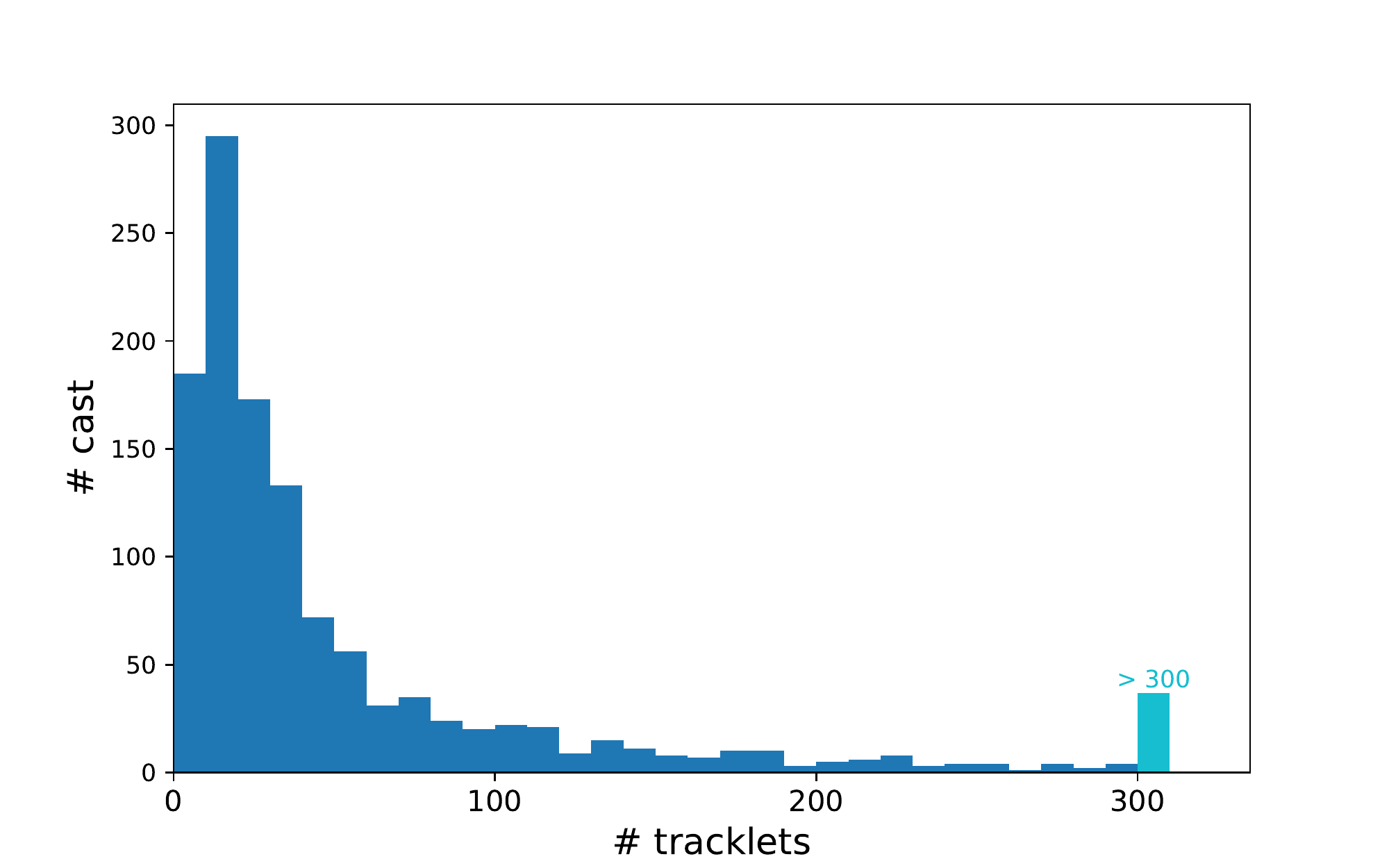}}
	\subfloat[\label{fig:statistic-4}]{\includegraphics[width=0.32\linewidth]{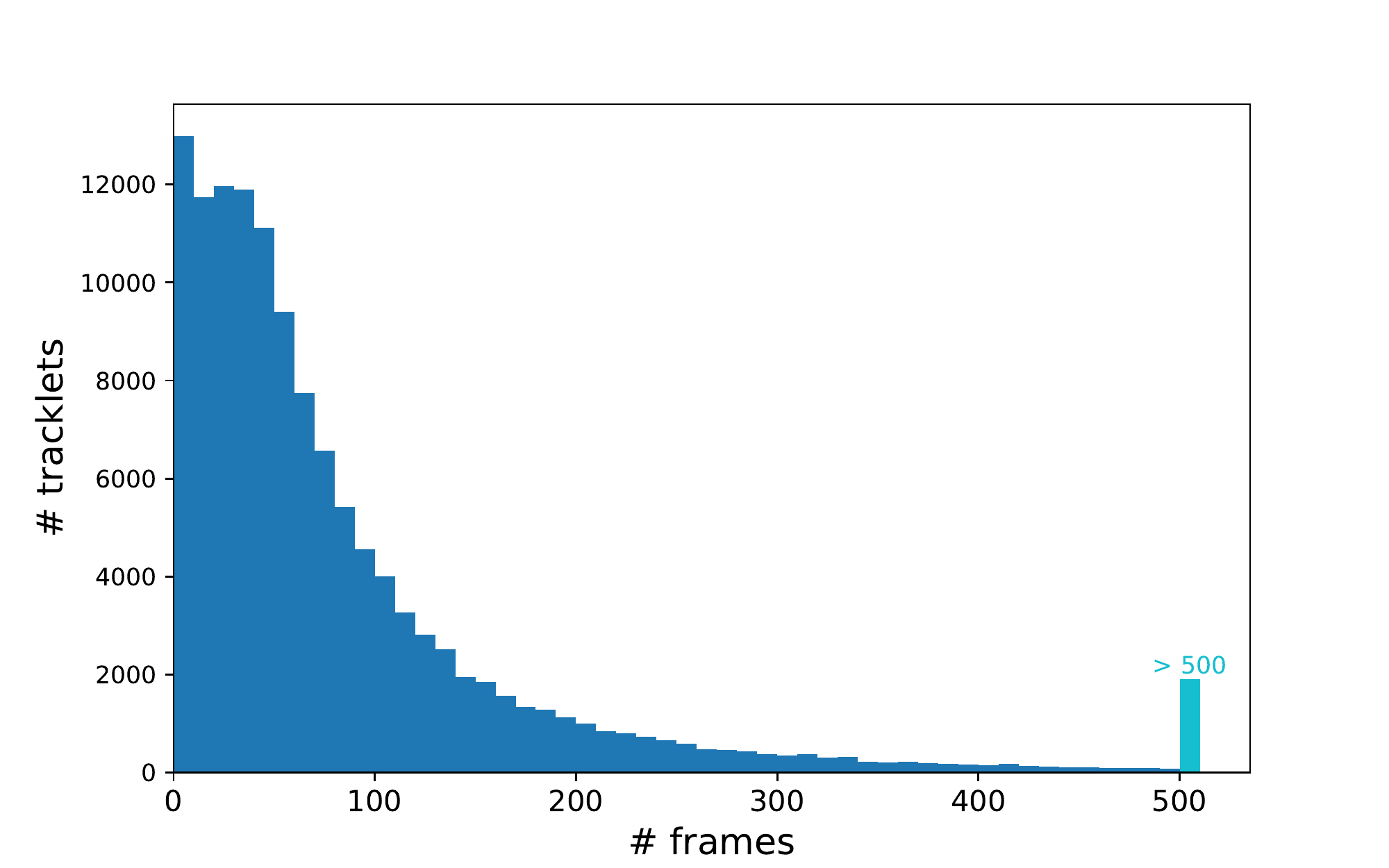}}
	\subfloat[\label{fig:statistic-5}]{\includegraphics[width=0.32\linewidth]{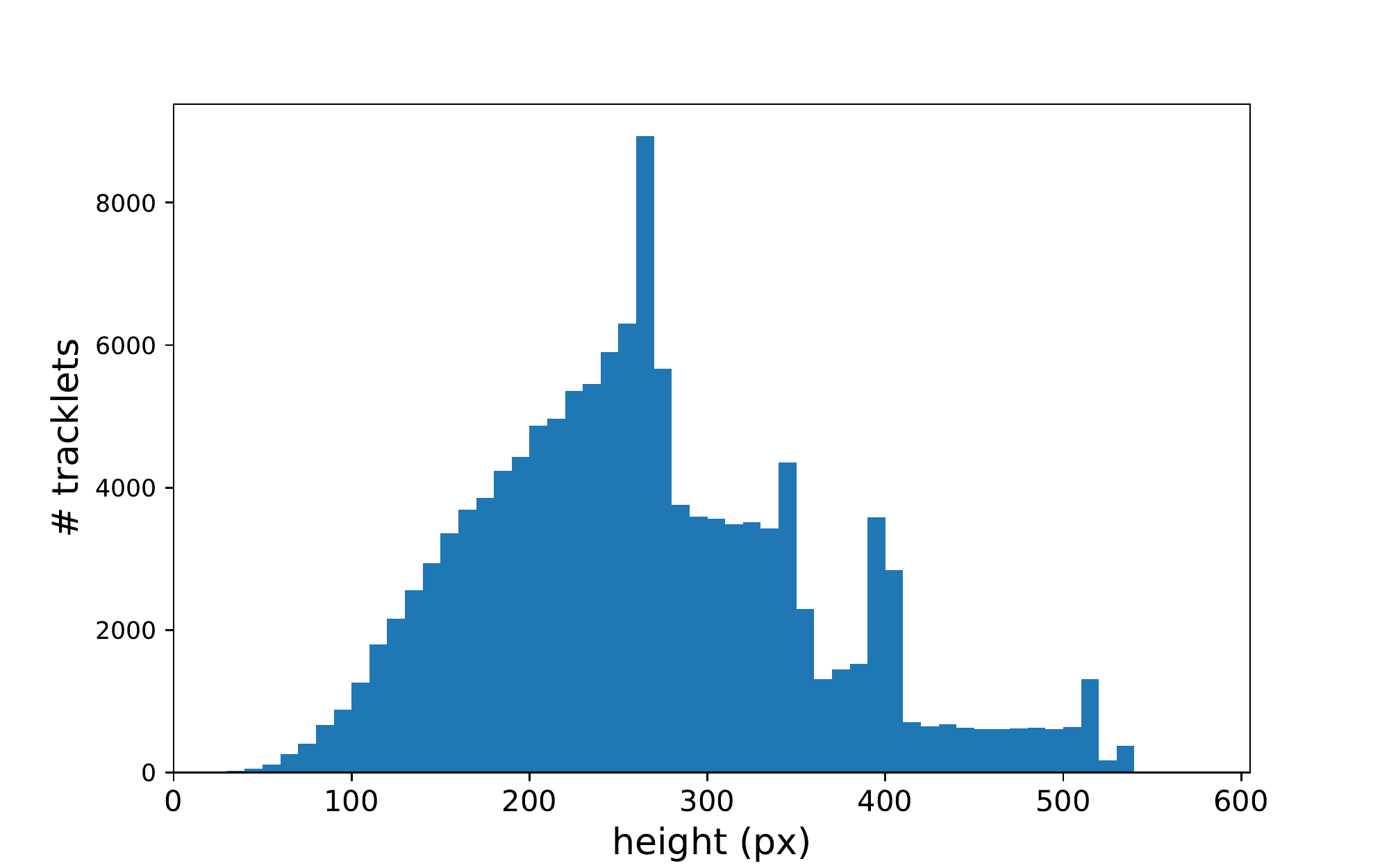}}
	\caption{
		Statistics of CSM dataset.
		(a): the tracklet number distribution over movies.
		(b): the tracklet number of each movie, both credited cast and ``others''.
		(c): the distribution of tracklet number over cast.
		(d): the distribution of length (frames) over tracklets.
		(e): the distribution of height (px) over tracklets.
	}
	\label{fig:dataset-statistic}
\end{figure}

\paragraph{\bf Query Set.}

For each movie in \emph{CSM},
we acquired the cast list from IMDB.
For those movies with more than $10$ cast,
we only keep the top $10$ according to the IMDB order,
which can cover the main characters for most of the movies.
In total, we obtained $1,218$ cast, which we refer to as the \emph{credited cast}.
For each credited cast,
we download a portrait from either its IMDB or TMDB homepage,
which will serve as the query portraits in \emph{CSM}.

\paragraph{\bf Gallery Set.}

We obtained the tracklets in the gallery set through five steps:

\begin{enumerate}
\item \emph{Detecting shots.}
A movie is composed of a sequence of shots.
Given a movie, we first detected the shot boundaries of the movies using
a fast shot segmentation technique~\cite{apostolidis2014fast,sidiropoulos2011temporal},
resulting in totally $200K$ shots for all movies.
For each shot, we selected $3$ frames as the \emph{keyframes}.

\item \emph{Annotating bounding boxes on keyframes.}
We then \emph{manually} annotated the person bounding boxes on keyframes
and obtained around $700K$ bounding boxes.

\item \emph{Training a person detector.}
We trained a person detector with the annotated bounding boxes.
Specifically, all the keyframes are partitioned into a training set
and a testing set by a ratio $7:3$.
We then finetuned a Faster-RCNN~\cite{ren2015faster} pre-trained on
MSCOCO~\cite{lin2014microsoft} on the training set.
On the testing set, the detector gets around $91\%$ mAP,
which is good enough for tracklet generation.

\item \emph{Generating tracklets.}
With the person detector as described above,
we performed per-frame person detection over all the frames.
By concatenating the bounding boxes across frames with
$\text{IoU} > 0.7$ \emph{within each shot}, we obtained
$127K$ trackets from the $192$ movies.

\item \emph{Annotating identities.}
Finally, we manually annotated the identities of all the
tracklets. Particularly,
each tracklet is annotated as one of the credited cast or as ``others''.
Note that the identities of the tracklets in each movie are annotated
independently to ensure high annotation quality with a reasonable budget.
Hence, being labeled as ``others'' means that the tracklet does not belong to
any credited cast of the corresponding movie.
\end{enumerate}


\section{Methodology}
\label{sec:method}

\begin{figure}[t]
	\centering
	\includegraphics[width=\linewidth]{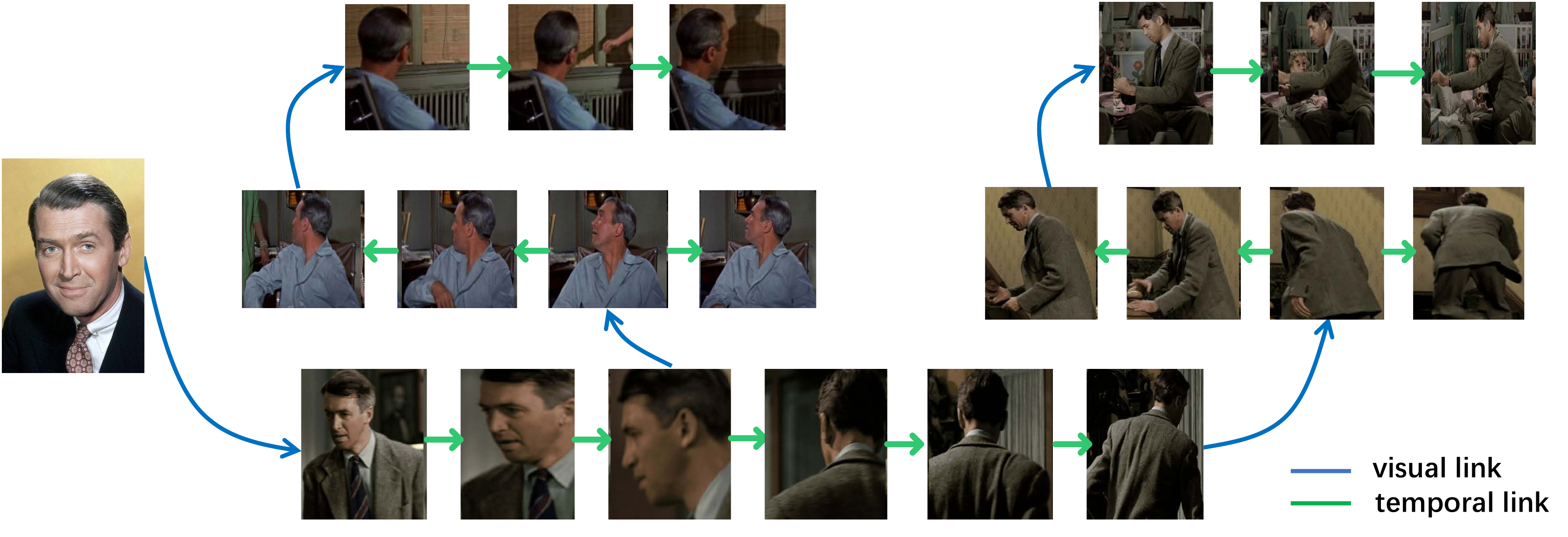}
	\caption{\small
		Visual links and temporal links in our graph.
		We only keep one strongest link for each pair of tracklets.
		And we can see that these two kinds of links are complementary.
		The former allows the identity information to be propagated
		among those instances that are similar in appearance, while the latter
		allows the propagation along a continuous tracklet, in which the instances can look significantly different. With both types of links incorporated, we can construct a more connected graph, which allows the identities to be propagated much further. 
	}
	\label{fig:graph}
\end{figure}


In this work, we aim to develop a method to find all the occurrences of
a person in a long video, \eg~a movie, with just a single portrait.
The challenge of this task lies in the vast gap of visual appearance
between the portrait (query) and the candidates in the gallery.


Our basic idea to tackle this problem by leveraging the inherent
\emph{identity invariance} along a person tracklet and propagate the identities
among instances via both visual and temporal links.
The visual and temporal links are complementary.
The use of both types of links allows
identities to be propagated much further than using either type alone.
However, how to propagate over a large, diverse, and noisy dataset
reliably remains a very challenging problem, considering that we only
begin with just a small number of labeled samples (the portraits).
The key to overcoming this difficulty is to be \emph{prudent}, only propagating
the information which we are certain about.
To this end, we propose a new propagation framework called
\emph{Progressive Propagation via Competitive Consensus}, which can effectively
identify confident labels in a competitive way.

\subsection{Graph Formulation}
\label{sub:graph}


The propagation is carried out over a graph among person instances.
Specifically, the propagation graph is constructed as follows.
Suppose there are $C$ cast in query set,
$M$ tracklets in gallery set,
and the length of $k$-th tracklet (denoted by $\tau_k$) is $n_k$, \ie~it contains $n_k$ instances.
The cast portraits and all the instances along the tracklets are treated
as graph nodes. Hence, the graph contains $N = C + \sum_{k=1}^M n_k$ nodes.
In particular, the identities of the $C$ cast portraits are known, and
the corresponding nodes are referred to as \emph{labeled nodes},
while the other nodes are called \emph{unlabled nodes}.


The propagation framework aims to propagate the identities
from the labeled nodes to the unlabeled nodes through both \emph{visual}
and \emph{temporal} links between them.
The \emph{visual links} are based on feature similarity.
For each instance (say the $i$-th), we can extract
a feature vector, denoted as $\vv_i$.
Each visual link is associated with an affinity value -- the affinity
between two instances $\vv_i$ and $\vv_j$ is defined to be their
cosine similarity as $w_{ij} = \vv_i^T \vv_j / (\|\vv_i\| \cdot \|\vv_j\|)$.
Generally, higher affinity value $w_{ij}$ indicates that $\vv_i$ and $\vv_j$
are more likely to be from the same identity.
The \emph{temporal links} capture the \emph{identity invariance} along
a tracklet, \ie~all instances along a tracklet should share the same identity.
In this framework, we treat the identity invariance as hard constraints,
which is enforced via a \emph{competitive consensus} mechanism.

For two tracklets with lengths $n_k$ and $n_l$, there can be $n_k \cdot n_l$
links between their nodes. Among all these links, the strongest link,
\ie~the one between the most similar pair, is the best to reflect the visual
similarity. Hence, we only keep one strongest link for each pair of tracklets
as shown in Figure~\ref{fig:graph},
which makes the propagation more reliable and efficient.
Also, thanks to the temporal links, such reduction would not compromise the
connectivity of the whole graph.


As illustrated in Figure~\ref{fig:graph}, the visual and temporal links are
complementary. The former allows the identity information to be propagated
among those instances that are similar in appearance, while the latter
allows the propagation along a continuous trajectory, in which the instances
can look significantly different.
With only visual links, we can obtain clusters in the feature space.
With only temporal links, we only have isolated tracklets.
However, with both types of links incorporated, we can construct a more
connected graph, which allows the identities to be propagated much further.

\subsection{Propagating via Competitive Consensus}

\begin{figure}[t]
	\centering
	\includegraphics[width=\linewidth]{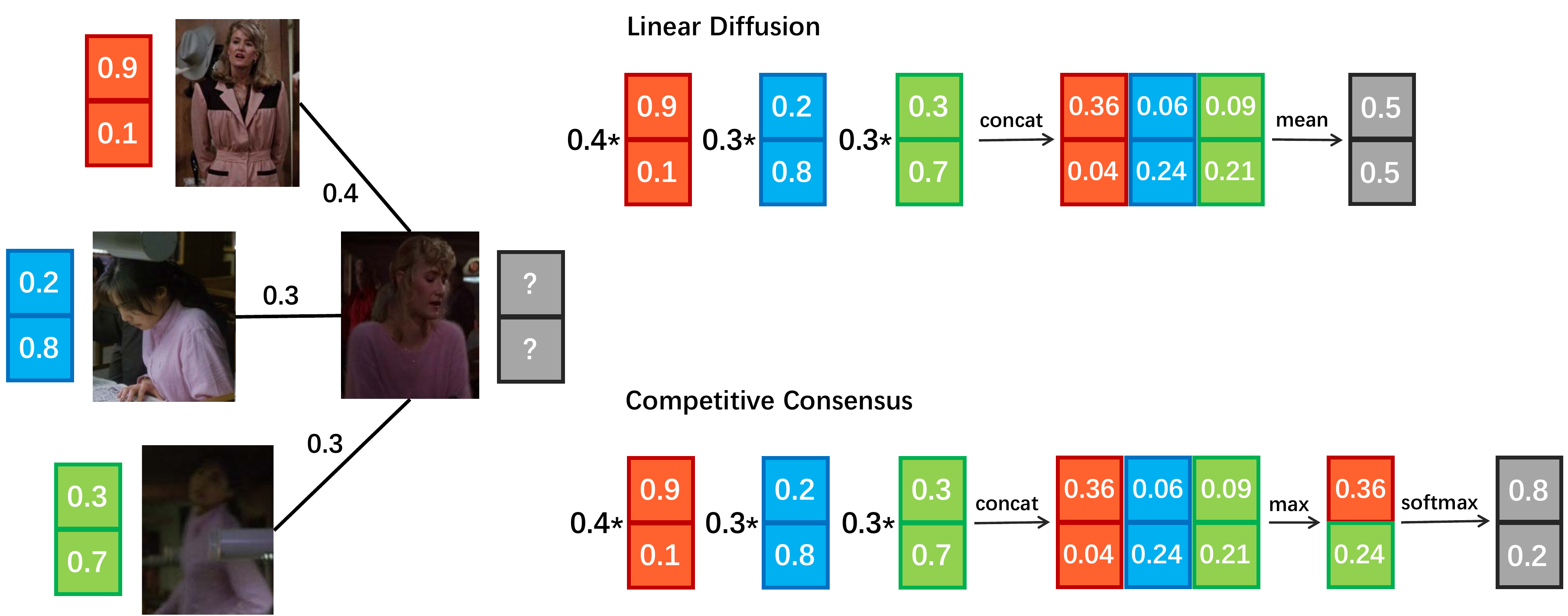}
	\caption{\small
		An example to show the difference between competitive consensus and linear diffusion.
		There are four nodes here and their probability vectors are shown by their sides. We are going to propagate labels from the left nodes to the right node.
		However, two of its neighbor nodes are noise.
		The calculation process of linear diffusion and competitive consensus are shown on the right side.
		We can see that in a graph with much noise,
		our competitive consensus,
		which aims to propagate the most confident information,
		is more robust.
	}
	\label{fig:cp}
\end{figure}

Each node of the graph is associated with a probability vector $\vp_i \in \mathbb{R}^C$,
which will be iteratively updated as the propagation proceeds.
To begin with, we set the probability vector for each labeled node to be
a one-hot vector indicating its label, and initialize all others to be zero
vectors.
Due to the identity invariance along tracklets, we enforce
all nodes along a tracklet $\tau_k$ to share the same probability vector,
denoted by $\vp_{\tau_k}$.
At each iteration, we traverse all tracklets and update
their associated probability vectors one by one.

\paragraph{\bf Linear Diffusion.}
Linear diffusion is the most widely used propagation scheme,
where a node would update its probability vector by taking a linear combination
of those from the neighbors.
In our setting with identity invariance,
the linear diffusion scheme can be expressed as follows:
\begin{equation}
\label{eq:linear-diffusion}
	\vp_{\tau_k}^{(t+1)} =
	\sum_{j \in \cN(\tau_k)} \alpha_{kj} \vp_j^{(t)}, \quad
	\text{ with }
	\alpha_{kj} = \frac{\tilde{w}_{kj}}{\sum_{j' \in \cN(\tau_k)} \tilde{w}_{kj'}}.
\end{equation}
Here, $\cN(\tau_k) = \cup_{i \in \tau_k} \cN_i$ is the set of all visual
neighbors of those instances in $\tau_k$.
Also, $\tilde{w}_{kj}$ is the \emph{affinity} of a neighbor node $j$ to the tracklet
$\tau_k$. Due to the constraint that there is only one visual link
between two tracklets (see Sec.~\ref{sub:graph}), each neighbor $j$ will
be connected to just one of the nodes in $\tau_k$, and $\tilde{w}_{kj}$ is set to
the affinity between the neighbor $j$ to that node.

However,
we found that the linear diffusion scheme yields poor performance in our experiments,
even far worse than the naive visual matching method.
An important reason for the poor performance is that errors will be mixed
into the updated probability vector and then propagated to other nodes.
This can cause catastrophic errors downstream, especially in a real-world
dataset that is filled with noise and challenging cases.

\paragraph{\bf Competitive Consensus.}
To tackle this problem, it is crucial to improve the reliability and
propagate the most confident information only.
Particularly, we should only trust those neighbors that provide strong
evidence instead of simply taking the weighted average of all neighbors.
Following this intuition,
we develop a novel scheme called \emph{competitive consensus}.

When updating $\vp_{\tau_k}$, the probability vector for the tracklet $\tau_k$,
we first collect the strongest evidence to support each identity $c$,
from all the neighbors in $\cN(\tau_k)$, as
\begin{equation}
  \eta_k(c) = \max_{j \in \cN(\tau_k)} \alpha_{kj} \cdot p_j^{(t)}(c),
\end{equation}
where the normalized coefficient $\alpha_{kj}$ is defined in
Eq.\eqref{eq:linear-diffusion}. Intuitively, an identity is \emph{strongly}
supported for $\tau_k$ if one of its neighbors assigns a high probability to it.
Next, we turn the evidences for individual identities into a probability vector
via a tempered softmax function as
\begin{equation}\label{eq:cc}
	p^{(t+1)}_{\tau_k}(c) = \exp(\eta_k(c)/T) / \sum_{c'=1}^C \exp(\eta_k(c')/T).
\end{equation}
Here, $T$ is a temperature the controls how much the probabilities concentrate
on the strongest identity.
In this scheme, all identities compete for getting high probability values in
$\vp^{(t+1)}_{\tau_k}$ by collecting the strongest supports from the neighbors.
This allows the strongest identity to stand out.

Competitive consensus can be considered as a coordinate ascent method to solve
Eq.~\ref{eq:gf}, where we introduce a binary variable $z_{kj}^{(c)}$ to
indicate whether the $j$-th neighbor is a trustable source for the class $c$
for the $k$-th tracklet. Here, $\mathcal{H}$ is the entropy. The constraint means that
one trustable source is selected for each class $c$ and tracklet $k$.
\begin{equation}
\label{eq:gf}
\max ~~ \sum_{c=1}^{C} p_{\tau_k}^{(c)} \sum_{j \in \cN(\tau_k)} \alpha_{kj} z_{kj}^{(c)} p_j^{(c)}  + \sum_{c=1}^{C}\mathcal{H}(p_{\tau_k}^{(c)})  ~~~~
s.t. \sum_{j \in \cN(\tau_k)} z_{kj}^{(c)} = 1.
\end{equation}

Figure~\ref{fig:cp} illustrates how linear diffusion and our competitive Consensus
work.
Experiments on CSM also show that competitive consensus significantly improves
the performance of the person search problem.

\subsection{Progressive Propagation}
\label{subsec:pp}

In conventional label propagation,
labels of all the nodes would be updated until convergence.
This way can be prohibitively expensive when the graph contains a large
number of nodes.
However, for the person search problem, this is unnecessary -- when we are
very confident about the identity of a certain instance, we don't have to
keep updating it.

Motivated by the analysis above,
we propose a \emph{progressive propagation} scheme to accelerate the
propagation process.
At each iteration,
we will fix the labels for a certain fraction of nodes that have the
highest confidence,
where the confidence is defined to be the maximum probability value
in $\vp_i$.
We found empirically that a simple freezing schedule, \eg~adding $10\%$ of
the instances to the label-frozen set, can already bring notable
benefits to the propagation process.

Note that the progressive scheme not only reduces computational cost
but also improves propagation accuracy.
The reason is that
without freezing, the noise and the uncertain nodes will keep affecting all the other nodes,
which can sometimes cause additional errors.
Experiments in~\ref{subsec:exp_csm} will show more details.


\section{Experiments}
\label{sec:exp}

\subsection{Evaluation protocol and metrics of CSM}
\label{subsec:exp_proto}

\begin{table}[t]
	\centering
	\begin{minipage}{.55\linewidth}
		\caption{train/val/test splits of CSM}
		\centering
		\label{tab:dataset-splits}
		\begin{tabular}{l|c|c|c|c}
			\hline
			& movies & cast  & tracklets & credited tracklets \\ \hline \hline
			train & 115    & 739   & 79K       & 47K                \\
			val   & 19     & 147   & 15K       & 8K                 \\
			test  & 58     & 332   & 32K       & 18K                \\ \hline
			total & 192    & 1,218 & 127K      & 73K                \\ \hline
		\end{tabular}
	\end{minipage}
	\begin{minipage}{.4\linewidth}
		\caption{query/gallery size}
		\centering
		\label{tab:dataset-qgsize}
		\begin{tabular}{c|c|c}
			\hline
			setting                                                   & ~query~ & ~gallery~ \\ \hline \hline
			\begin{tabular}[c]{@{}c@{}}IN \\ \small{(per movie)}\end{tabular} & 6.4   & 560.5   \\ \hline
			CROSS                                                     & 332   & 17,927  \\ \hline
		\end{tabular}
	\end{minipage}
\end{table}

The $192$ movies in CSM are partitioned into training (train), validation (val) and testing (test) sets.
Statistics of these sets are shown in Table~\ref{tab:dataset-splits}.
Note that we make sure that there is no overlap between the cast of different sets.
\ie~the cast in the testing set would not appear in training and validation.
This ensures the reliability of the testing results.

Under the Person Search with One Portrait setting,
one should rank all the tracklets in the gallery given a query.
For this task, we use \emph{mean Average Precision (mAP)} as the evaluation metric.
We also report the recall of tracklet identification results in our experiments
in terms of R@k.
Here, we rank the identities for each tracklet according to
their probabilities.
R@k means the fraction of tracklets for which the correct identity is
listed within the top $k$ results.

We consider two test settings in the CSM benchmark named
``search cast in a movie'' (IN) and ``search cast across all movies'' (ACROSS).
The setting ``IN'' means the gallery consists of just the tracklets from one movie,
including the tracklets of the credited cast and those of ``others''.
While in the ``ACROSS'' setting, the gallery comprises all the tracklets of credited cast in testing set.
Here we exclude the tracklets of ``others'' in the ``ACROSS'' setting
because ``others'' just means that it does not belong to any one of the credited cast of a particular movie rather than all the movies in the dataset
as we have mentioned in Sec.~\ref{sec:dataset}.
Table~\ref{tab:dataset-qgsize} shows the query/gallery sizes of each setting.

\subsection{Implementation Details}
\label{subsec:exp_imp}

We use two kinds of visual features in our experiments.
The first one is the IDE feature~\cite{zheng2016mars} widely used in person re-id.
The IDE descriptor is a CNN feature of the whole person instance,
extracted by a Resnet-50~\cite{he2016deep}, which is pre-trained on ImageNet~\cite{russakovsky2015imagenet}
and finetuned on the training set of CSM.
The second one is the \emph{face feature},
extracted by a Resnet-101, which is trained on MS-Celeb-1M~\cite{guo2016ms}.
For each instance, we extract its IDE feature and the face feature of the face region, which is detected by a face detector~\cite{zhang2016joint}.
All the visual similarities in experiments are calculated by cosines similarity between the visual features.

\begin{table}[]
	\centering
	\caption{Results on CSM under Two Test Settings}
	\begin{tabular}{l|c|ccc|c|ccc}
		\hline
		& \multicolumn{4}{c|}{IN}       & \multicolumn{4}{c}{ACROSS}   \\ \hline
		& ~~mAP~~   & ~~R@1~~   & ~~R@3~~   & ~~R@5~~   & ~~mAP~~   & ~~R@1~~   & ~~R@3~~   & ~~R@5~~   \\ \hline\hline
		FACE     & 53.33 & 76.19 & 91.11 & 96.34 & 42.16 & 53.15 & 61.12 & 64.33 \\
		IDE      & 17.17 & 35.89 & 72.05 & 88.05 & 1.67  & 1.68  & 4.46  & 6.85  \\
		FACE+IDE & 53.71 & 74.99 & 90.30 & 96.08 & 40.43 & 49.04 & 58.16 & 62.10 \\
		LP       & 8.19  & 39.70 & 70.11 & 87.34 & 0.37  &  0.41 &  1.60 & 5.04        \\ \hline
		PPCC-v   & 62.37 & \textbf{84.31} & \textbf{94.89} & \textbf{98.03} & 59.58 & \textbf{63.26} & \textbf{74.89} & \textbf{78.88}   \\
		PPCC-vt  & \textbf{63.49} & 83.44 & 94.40 & 97.92 & \textbf{62.27} & 62.54  &  73.86     &  77.44     \\ \hline
	\end{tabular}
	\label{tab:exp-csm}
\end{table}

\subsection{Results on CSM}
\label{subsec:exp_csm}

We set up four baselines for comparison:
\textbf{(1) FACE:} To match the portrait with the tracklet in the gallery by face feature similarity.
Here we use the mean feature of all the instances in the tracklet to represent it.
\textbf{(2) IDE:} Similar to FACE, except that the IDE features are used rather than the face features.
\textbf{(3) IDE+FACE:} To combine face similarity and IDE similarity for matching, respectively
with weights $0.8$ and $0.2$.
\textbf{(4) LP:} Conventional label propagation with linear diffusion with both visual and temporal links.
Specifically, we use face similarity as the visual links between portraits and
candidates and the IDE similarity as the visual links between different candidates.
We also consider two settings of the proposed
Progressive Propagation via Competitive Consensus method.
\textbf{(5) PPCC-v:} using only visual links.
\textbf{(6) PPCC-vt: } the full config with both visual and temporal links.

From the results in Table~\ref{tab:exp-csm}, we can see that:
(1) Even with a very powerful CNN trained on a large-scale dataset,
matching portrait and candidates by visual cues cannot solve the person search problem well due to the big gap of visual appearances between the portraits and the candidates.
Although face features are generally more stable than IDE features, they
would fail when the faces are invisible,
which is very common in real-world videos like movies.
(2) Label propagation with linear diffusion gets very poor results,
even worse than the matching-based methods.
(3) Our approach raises the performance by a considerable margin.
Particularly,
the performance gain is especially remarkable on the more challenging ``ACROSS'' setting
($62.27$ with ours vs. $42.16$ with the visual matching method).

\begin{figure}[t]
	\centering
	\subfloat[Under ``IN'' setting]{
		\includegraphics[width=0.45\linewidth]{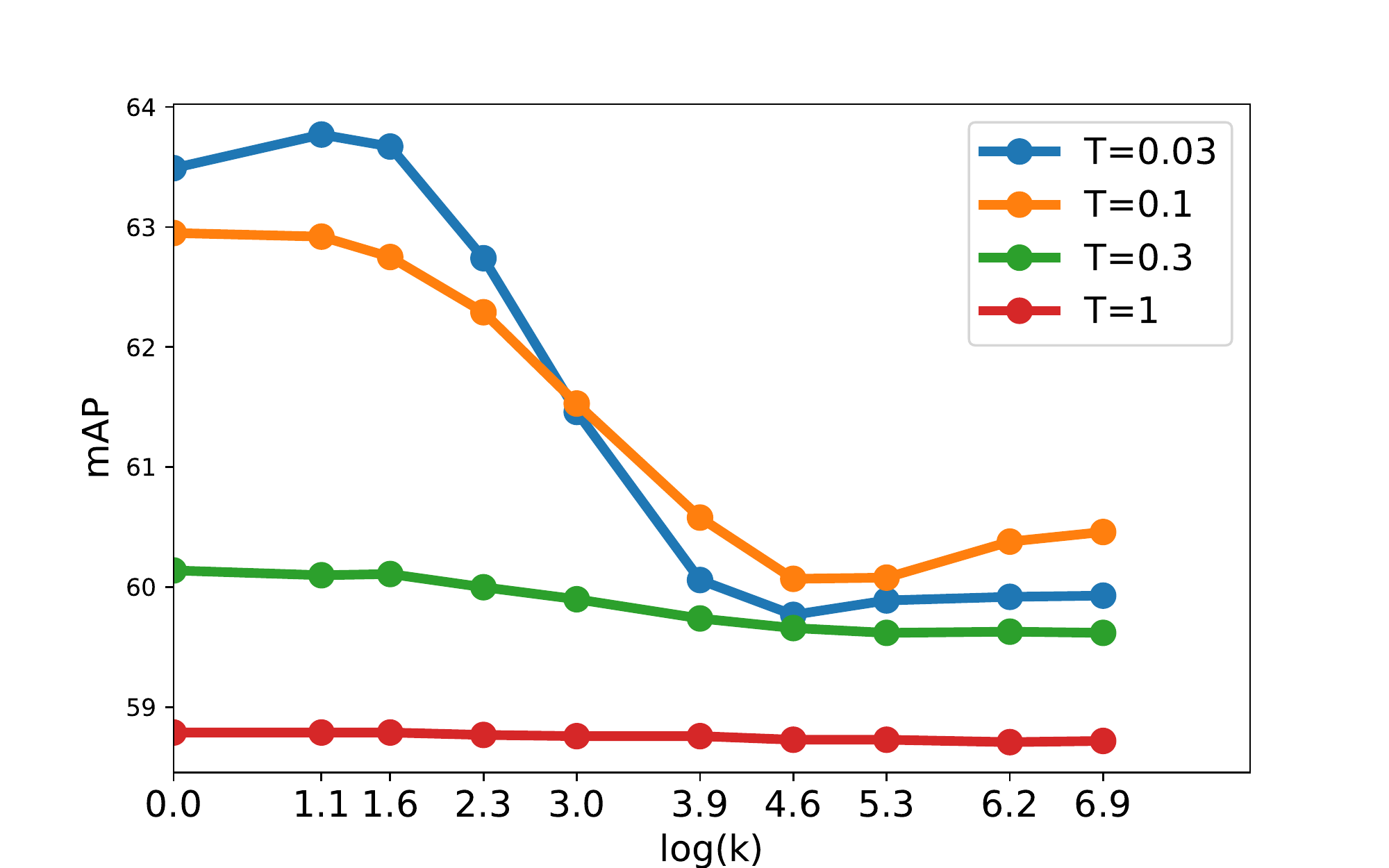}
		\label{fig:exp-cc-1}
	} \hfill
	\subfloat[Under ``ACROSS'' setting]{
		\includegraphics[width=0.45\linewidth]{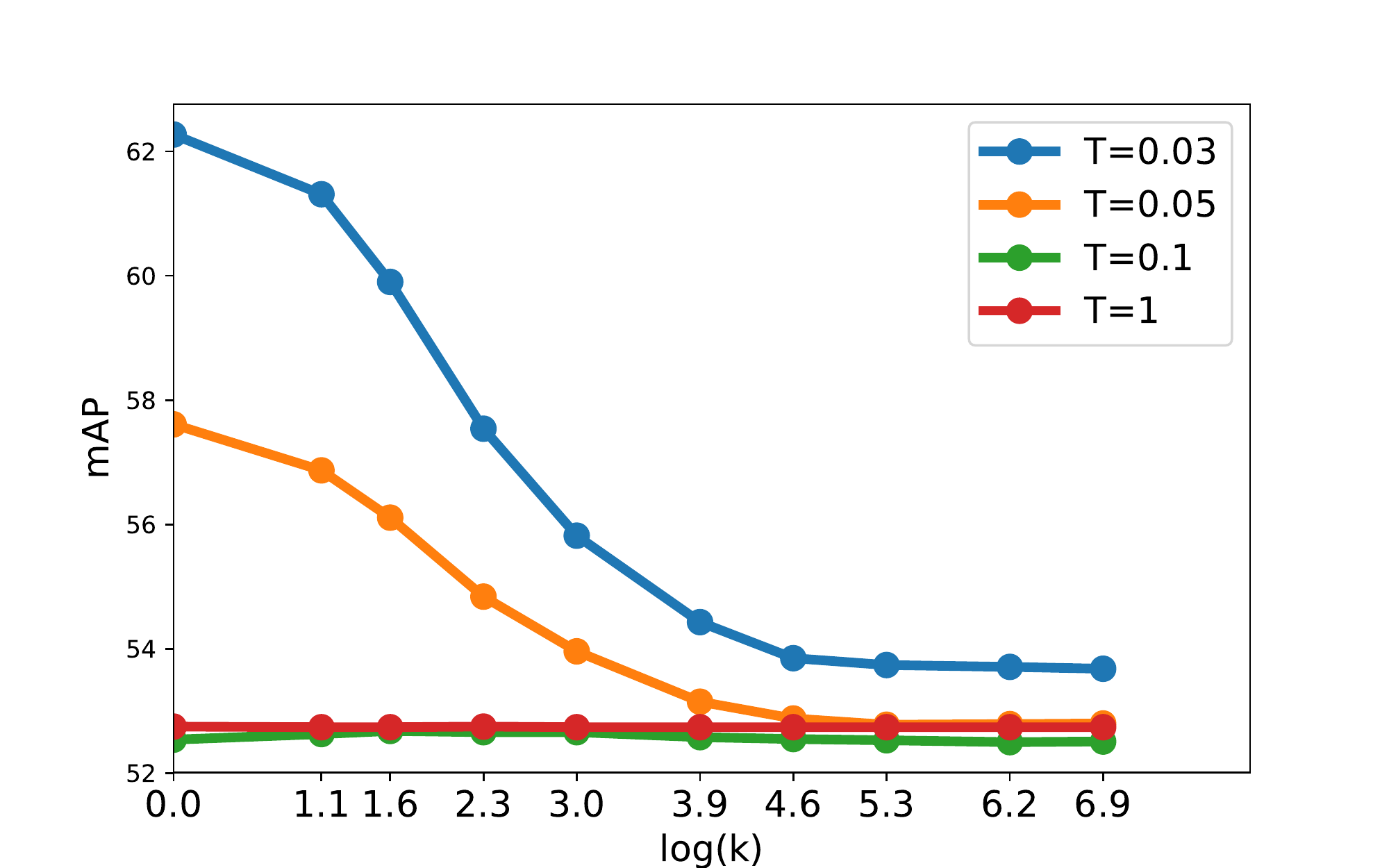}
		\label{fig:exp-cc-2}
	}
	\vspace{-5pt}
	\caption{\footnotesize
		mAP of different settings of competitive consensus.
		Comparison between different temperatures(T) of softmax
		and different settings of $k$ (in top-$k$ average).
	}
	\label{fig:exp-cc}
\end{figure}

\paragraph{\bf Analysis on Competitive Consensus}.
To show the effectiveness of \emph{Competitive Consensus},
we study different settings of the Competitive Consensus scheme
in two aspects:
(1) The $\max$ in Eq.~\eqref{eq:cc} can be relaxed to top-$k$ average.
Here $k$ indicates the number of neighbors to receive information from.
When $k=1$, it reduces to only taking the maximum, which is what we use in PPCC.
Performances obtained with different $k$ are shown in Fig.~\ref{fig:exp-cc}.
(2) We also study on the ``softmax'' in~ Eq.\eqref{eq:cc} and compare results
between different temperatures of it. The results are also shown in Fig.~\ref{fig:exp-cc}.
Clearly, using smaller temperature of softmax significantly boosts the performance.
This study supports what we have claimed when designing \emph{Competitive Consensus}:
we should only propagate the most confident information in this task.

\paragraph{\bf Analysis on Progressive Propagation}.
Here we show the comparison between our progressive updating scheme and the conventional scheme that updates all the nodes at each iteration.
For progressive propagation,
we try two kinds of freezing mechanisms:
(1) \emph{Step} scheme means that we set the freezing ratio of each iteration and the ratio are raised step by step.
More specifically, the freezing ratio $r$ is set to $r = 0.5 + 0.1 \times \text{iter}$ in our experiment.
(2) \emph{Threshold} scheme means that we set a threshold,
and each time we freeze the nodes whose max probability to a particular identity is greater than the threshold.
In our experiments, the threshold is set to $0.5$.
The results are shown in Table~\ref{tab:exp-progressive},
from which we can see the effectiveness of the progressives scheme.

\begin{table}[t]
	\centering
	\caption{Results of Different Updating Schemes}
	\begin{tabular}{l|c|ccc|c|ccc}
		\hline
		& \multicolumn{4}{c|}{IN}       & \multicolumn{4}{c}{ACROSS}   \\ \hline
		& ~~mAP~~   & ~~R@1~~   & ~~R@3~~   & ~~R@5~~   & ~~mAP~~   & ~~R@1~~   & ~~R@3~~   & ~~R@5~~   \\ \hline\hline
		Conventional     & 60.54 & 76.64 & 91.63 & 96.70 & 57.42 & 54.60 & 63.31 & 66.41 \\
		Threshold      & 62.51 & 81.04 & 93.61 & 97.48 & 61.20  & 61.54  & 72.31  & 76.01  \\
		Step  & \textbf{63.49} & \textbf{83.44} & \textbf{94.40} & \textbf{97.92} & \textbf{62.27} & \textbf{62.54}  &  \textbf{73.86}     &  \textbf{77.44}     \\ \hline
	\end{tabular}
	\label{tab:exp-progressive}
\end{table}

\begin{figure}[t]
	\centering
	\includegraphics[width=\linewidth]{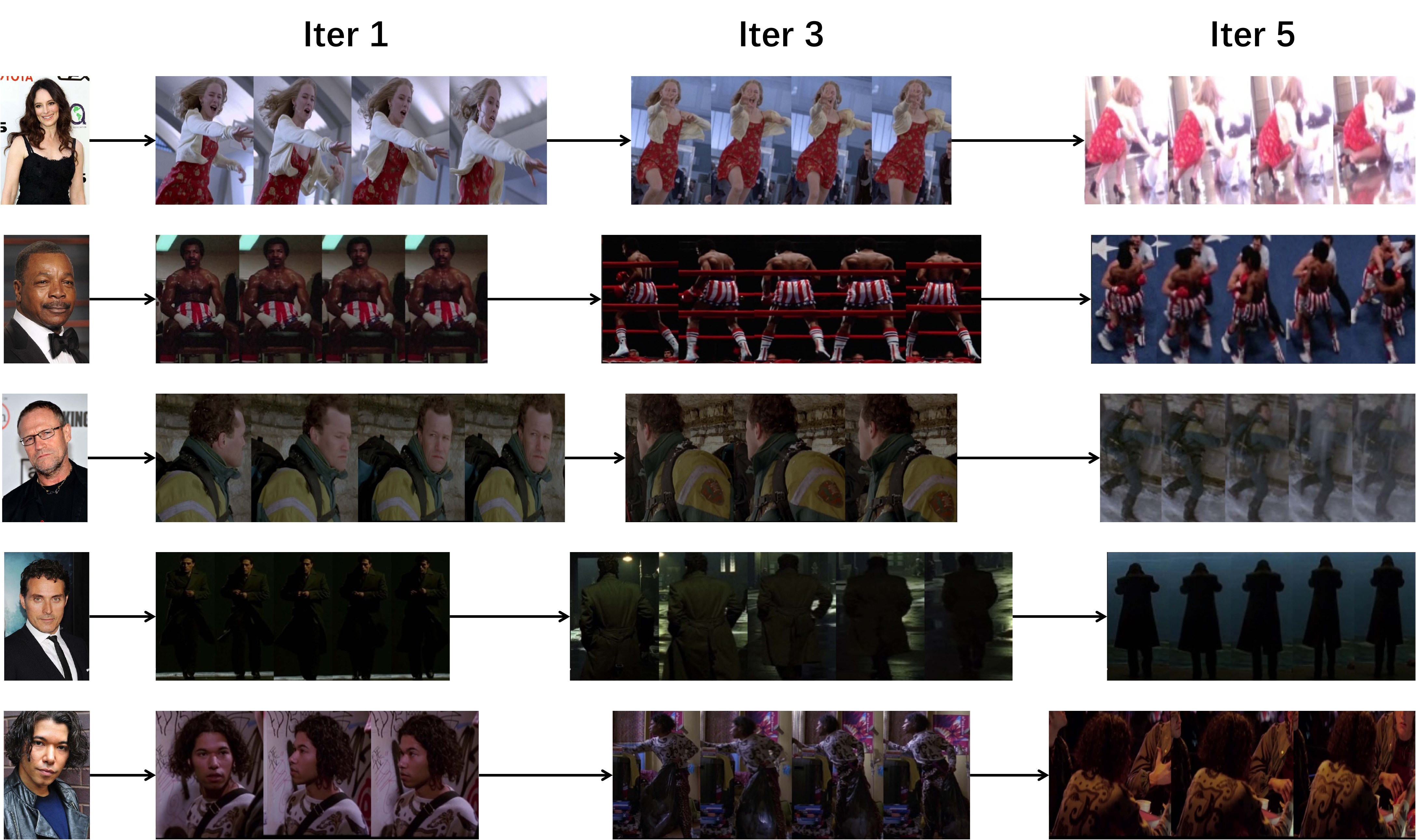}
	\caption{
		Some samples that are correctly searched in different iterations.
		}
		\label{fig:cases}
		\end{figure}

\paragraph{\bf Case Study}.
We show some samples that are correctly searched in different iterations
in Fig.~\ref{fig:cases}.
We can see that the easy cases,
which are usually with clear frontal faces,
can be identified at the beginning.
And after iterative propagation,
the information can be propagated to the harder samples.
At the end of the propagation,
even some very hard samples,
which are non-frontal, blurred, occluded and under extreme illumination,
can be propagated a right identity.

\section{Conclusion}
\label{sec:conclusion}

In this paper,
we studied a new problem named \emph{Person Search in Videos with One Protrait},
which is challenging but practical in the real world.
To promote the research on this problem,
we construct a large-scale dataset \emph{CSM},
which contains $127K$ tracklets of $1,218$ cast from $192$ movies.
To tackle this problem, we proposed a new
framework that incorporates both visual and temporal links for identity propagation,
with a novel \emph{Progressive Propagation vis Competitive Consensus} scheme.
Both quantitative and qualitative studies show
the challenges of the problem
and the effectiveness of our approach.


\section{Acknowledgement}
\label{sec:acknowledgement}

This work is partially supported by the Big Data Collaboration Research grant from SenseTime Group (CUHK Agreement No. TS1610626), the General Research Fund (GRF) of Hong Kong (No. 14236516).

\bibliographystyle{splncs04}
\bibliography{egbib}
\end{document}